\documentclass[lettersize,journal]{IEEEtran}
\usepackage{standalone} 
\usepackage{tikz}
\usepackage{xcolor}
\usepackage{xstring} 
\usepackage{amsmath,amsfonts}
\usepackage{algorithmic}
\usepackage{algorithm}
\usepackage{array}
\usepackage{colortbl}
\usepackage[caption=false,font=normalsize,labelfont=sf,textfont=sf]{subfig}
\usepackage{textcomp}
\usepackage{stfloats}
\usepackage{url}
\usepackage{verbatim}
\usepackage{graphicx}
\usepackage{cite}
\usepackage{amsmath}
\usepackage{booktabs}
\usepackage{arydshln} 
\usepackage{tabularx}   
\usepackage{pgf}
\usepackage{multirow}
\usepackage{cuted}
\usepackage{microtype}
\usepackage{subfig}
\usepackage{etoolbox}
\usepackage{ifthen} 
\usepackage{placeins}

\usepackage{siunitx} 
\usepackage{ragged2e} 

\usepackage[hidelinks]{hyperref}  
\usepackage{cleveref}
\usepackage[acronym]{glossaries}
\glsdisablehyper
\definecolor{matplotlibgreen}{HTML}{4CAF50} 

\makeatletter
\newcommand{\conditionalresizebox}[1]{%
  \@ifclasswith{IEEEtran}{draftcls}{%
    #1%
  }{%
    \resizebox{\columnwidth}{!}{#1}%
  }%
}
\makeatother

\newcolumntype{H}{>{\columncolor{yellow!10}}c} 

\newboolean{arxivflag}
\setboolean{arxivflag}{false}  

\newcommand{\checkarxivfile}{%
    \IfFileExists{is_arxiv_sub.flag}{%
        \setboolean{arxivflag}{true}%
    }{%
    }%
}

\AtBeginDocument{\checkarxivfile}

\begin{document}

\newacronym{rd}{RD}{rate-distortion}
\newacronym{msssim}{MS-SSIM}{multi-scale structural similarity index measure \cite{msssim_Wang_2003}}
\newacronym{psnr}{PSNR}{peak signal-to-noise ratio}
\newacronym{mse}{MSE}{mean squared error}
\newacronym{bpp}{bpp}{bits per pixel}
\newacronym{gmac}{GMAC}{billion multiply-accumulate operations}
\newacronym{mp}{MP}{megapixel}
\newacronym{iou}{IoU}{intersection over union}
\newacronym[longplural={convolutional neural networks \cite{cnn_Lecun_1998}}]{cnn}{CNN}{convolutional neural network \cite{cnn_Lecun_1998}}
\newacronym{cdf}{CDF}{cumulative distribution function}
\newacronym[\glsshortpluralkey={s},\glslongpluralkey={seconds}]{s}{s}{second}
\newacronym{cpm}{CPM}{cumulative probability model}
\newacronym[longplural={autoencoders \cite{autoencoder_Kramer_1991}}]{ae}{AE}{autoencoder \cite{autoencoder_Kramer_1991}}
\newacronym{isp}{ISP}{image signal processor}  
\newacronym[longplural={hyperpriors \cite{hyperprior_Balle_2018}}]{hp}{HP}{hyperprior \cite{hyperprior_Balle_2018}}
\newacronym{lut}{LUT}{look-up table}
\newacronym{cpu}{CPU}{central processing unit}
\newacronym{jdc}{JDC}{joint denoising and compression}  
\newacronym{jddc}{JDDC}{joint denoising, demosaicing and compression}

\newacronym{xmp}{XMP}{Extensible Metadata Platform (ISO 16684)}  
\newacronym{rgb}{RGB}{red green blue}  
\newacronym{cfa}{CFA}{color filter array}
\newacronym{ss}{s.s.}{shutter speed}
\newacronym{nind}{NIND}{Natural Image Noise Dataset \cite{nind_Brummer_2019}}
\newacronym{rawnind}{RawNIND}{Raw Natural Image Noise Dataset}
\newacronym{sidd}{SIDD}{Smartphone Image Noise Dataset \cite{sidd_Abdelhamed_2018}}
\newacronym{cfp}{COM:FP}{Wikimedia Commons Featured Pictures \cite{fp_Wikimedia_2020}}
\newacronym{fp}{FP}{floating-point}
\newacronym{bm3d}{BM3D}{block-matching and 3D filtering \cite{bm3d_Dabov_2006}}
\newacronym{nlm}{NLM}{non-local means \cite{nlm_Buades_2005}}

\newacronym{sid}{SID}{See-in-the-Dark \cite{SID_Chen_2018} dataset}
\newacronym{rec2020}{Rec. 2020}{ITU-R Recommendation BT.2020 \cite{rec2020} color space}
\newacronym{srgb}{sRGB}{standard RGB color space \cite{srgb} (IEC 61966-2-1:1999)}  
\newacronym{isostd}{ISO}{International Organization for Standardization}  
\newacronym{isocam}{ISO}{ISO \textit{camera sensitivity} standard}  
\newacronym{relu}{ReLU}{Rectified Linear Unit \cite{relu_Nair_2010}}
\newacronym{xyz}{XYZ}{CIE 1931 XYZ color space}
\newacronym{camrgb}{CamRGB}{camera-specific color space}
\newacronym{nn}{NN}{neural network}
\newacronym{lambda}{$\lambda$}{rate-distortion training parameter increasing the image quality (thus bitrate) of compression models}

\title{Learning Joint Denoising, Demosaicing, and Compression from the Raw Natural Image Noise Dataset}

\author{
  \IEEEauthorblockN{Benoit Brummer and Christophe De Vleeschouwer}\\
  \IEEEauthorblockA{University of Louvain, Louvain-la-Neuve, Belgium}
}




\maketitle

\begin{abstract}

This paper introduces the \gls{rawnind}, a diverse collection of paired raw images designed to support the development of denoising models that generalize across sensors, image development workflows, and styles. Two denoising methods are proposed: one operates directly on raw Bayer data, leveraging computational efficiency, while the other processes linear RGB images for improved generalization to different sensors, with both preserving flexibility for subsequent development. Both methods outperform traditional approaches which rely on developed images. Additionally, the integration of denoising and compression at the raw data level significantly enhances rate-distortion performance and computational efficiency. These findings suggest a paradigm shift toward raw data workflows for efficient and flexible image processing.

\end{abstract}

\begin{IEEEkeywords}
Image~compression, image~denoising, image~processing, raw~images, dataset, photography
\end{IEEEkeywords}

\begin{figure*}[!t]
	\centering
	\ifthenelse{\boolean{arxivflag}}
	{\includegraphics[width=\textwidth]{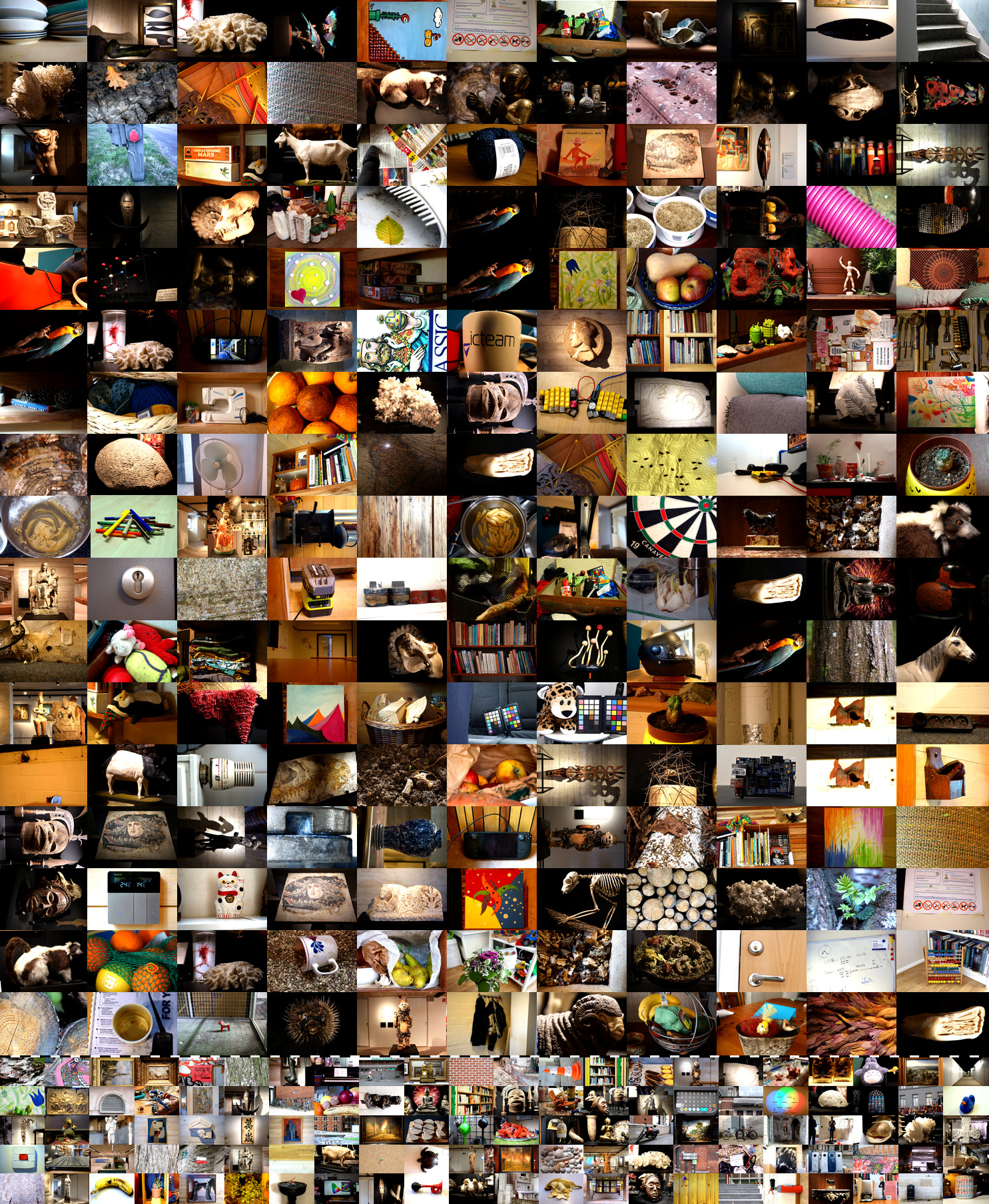}}
	{\includegraphics[width=\textwidth]{figures/combined_vertical_mosaic_q80.jpg}}
	\caption{Preview of the \gls{rawnind}'s Bayer (top) and X-Trans (bottom) ground-truth images}
	\label{fig:mosaic}
\end{figure*}

\section{Introduction}

Image denoising and compression are fundamental tasks in digital photography and image development, but current methods applied to developed images suffer from limitations in performance and generalization. These limitations arise because traditional compression methods and most of the literature in image denoising focus on developed images, which have undergone complex and often proprietary transformations applied by a camera \gls{isp} or an image development software. This work presents a methodology that leverages raw sensor data and the \acrfull{rawnind}\glsunset{rawnind}, a contribution of this work, to achieve significant improvements in computational efficiency (four to sixteen times fewer operations), produce high-quality images from noisy raw data, and effectively generalize across diverse camera sensors and development workflows.

In contrast, raw sensor data offers a more consistent and flexible representation of the captured scene. Unlike developed images, raw data preserves the linear sensor output and avoids artifacts introduced during image development, making it more suitable for denoising and compression tasks. However, raw image manipulation introduces its own challenges, as each sensor has unique characteristics, such as noise patterns and native color spaces. These variations have historically led to the assumption that models trained on one camera's raw data would not generalize to others.

To address these challenges, we introduce the \textit{\acrfull{rawnind}}, a collection of paired raw images from a variety of camera sensors. \Acrshort{rawnind} enables the development of models that generalize across sensors, workflows, and noise conditions. Using this dataset, we explore two complementary denoising methods:
\begin{itemize}
\item \textbf{Raw Bayer Image Denoising:} This approach offers significant computational efficiency (reducing processing by a factor of four) by denoising raw Bayer data directly, operating on one-quarter of the spatial dimensions and deferring demosaicing until the final step.
\item \textbf{Linear \acrshort{rgb} Denoising:} This method operates on debayered images in a standardized linear \acrshort{rgb} color space, maintaining high image quality and facilitating better generalization across different \acrfullpl{cfa}\glsunset{cfa} while maintaining the advantages of raw data.
\end{itemize}

Both of the proposed denoising methods outperform approaches that take developed images as input, and they integrate seamlessly into any image development workflow, providing adaptability for any real-world application.

Building on these denoising advancements, we then integrate denoising and compression into a unified framework that operates directly on raw sensor data. This joint approach significantly enhances both computational efficiency at the encoder stage and \acrlong{rd} performance by addressing noise at the earliest stage of the image pipeline. Unlike conventional workflows which compress fully developed images, our method compresses raw data, preserving flexibility for subsequent editing. It also enables new compression paradigms where raw images are stored alongside lightweight \acrshort{xmp} sidecar files, which contain all the necessary development instructions to produce the final image.

This paper makes the following key contributions:
\begin{itemize}
\item \textbf{\acrshort{rawnind} Dataset:} A novel dataset of paired raw images enabling generalized denoising and compression across diverse sensors and noise conditions.
\item \textbf{Denoising Methods:} Two high-performing approaches for raw image denoising—considering Bayer denoising for computational efficiency and linear \acrshort{rgb} representations for different \acrshortpl{cfa}s—that generalize across all image development workflows.
\item \textbf{Integrated Denoising and Compression:} A joint model that significantly improves \acrlong{rd} performance and computational efficiency compared to traditional sequential pipelines.
\item \textbf{Compression Paradigm:} We propose a paradigm for storing compressed raw images with metadata (\acrshort{xmp} sidecar files containing development instructions), enabling non-destructive editing and efficient storage by including instructions to convert the raw data into a developed image.
\end{itemize}

The rest of the paper is organized as follows: \Cref{sec:related_work} reviews related work on image denoising and compression. \Cref{sec:approach} describes our proposed methods and the creation of RawNIND. \Cref{sec:experiments} details the experimental setup and evaluation metrics, while \Cref{sec:results} presents the results and discusses key findings. Finally, \Cref{sec:conclusion} concludes the paper and outlines directions for future research.

\section{Related Work}
\label{sec:related_work}

\subsection{Image Denoising}

Image denoising aims to suppress noise while preserving important image details. Traditional methods, such as \acrfull{nlm}\glsunset{nlm}, wavelet-based denoising \cite{wavelets_denoise_Donoho_1995}, and \acrshort{bm3d} \cite{bm3d_Dabov_2006}, rely on mathematical models to exploit redundancies in image structure. Among these, \acrshort{bm3d} remains a widely used benchmark due to its high-quality results under controlled conditions, despite its computational intensity, while \acrshort{nlm} and wavelet-based methods are implemented in raw image development software such as Darktable, due to their simplicity and efficiency.

The emergence of deep learning techniques has shifted the focus to data-driven approaches. \Acrfullpl{cnn}\glsunset{cnn}, including architectures like U-Net \cite{unet_Ronneberger_2015} and methods such as DnCNN \cite{learned_denoising_Zhang_2017}, and CBDNet \cite{learned_denoising_Guo_2019}, have shown significant improvements over traditional methods. However, these models often rely on synthetic noise for training and are sensitive to mismatches between training and testing noise distributions. While datasets like \acrshort{sid} \cite{SID_Chen_2018}, \acrshort{sidd} \cite{sidd_Abdelhamed_2018}, and the \acrfull{nind} provide real noisy-clean image pairs for training, their focus on specific sensors and development workflows limits their applicability to diverse real-world scenarios.

Critically, most existing work evaluates denoising on developed images, which does not reflect real-world use cases. In practice, denoising is almost always performed by the camera's \gls{isp} or raw development software, either directly on raw data or before applying non-linear transformations like sharpening. Post-development denoising is rare and generally suboptimal, as the introduction of non-linearities limits both denoising performance and its integration into the imaging pipeline. These limitations underscore the importance of addressing noise earlier in the processing workflow, such as directly on raw data.

Methods combining denoising and demosaicing \cite{demosaic_denoise_Hirakawa_2005, deep_demosaic_denoise_Gharbi_2016, demosaic_denoise_Liu_2020} have demonstrated improved noise suppression by leveraging raw data's linear characteristics with respect to illumination. However, these studies often assume a single camera model, leaving cross-sensor generalization largely unexplored. This work addresses this gap by explicitly evaluating generalization across multiple sensors and workflows using raw image data.

\subsection{Raw Image Development Pipelines}

\begin{figure*}
	\centering
	\includegraphics[width=\textwidth]{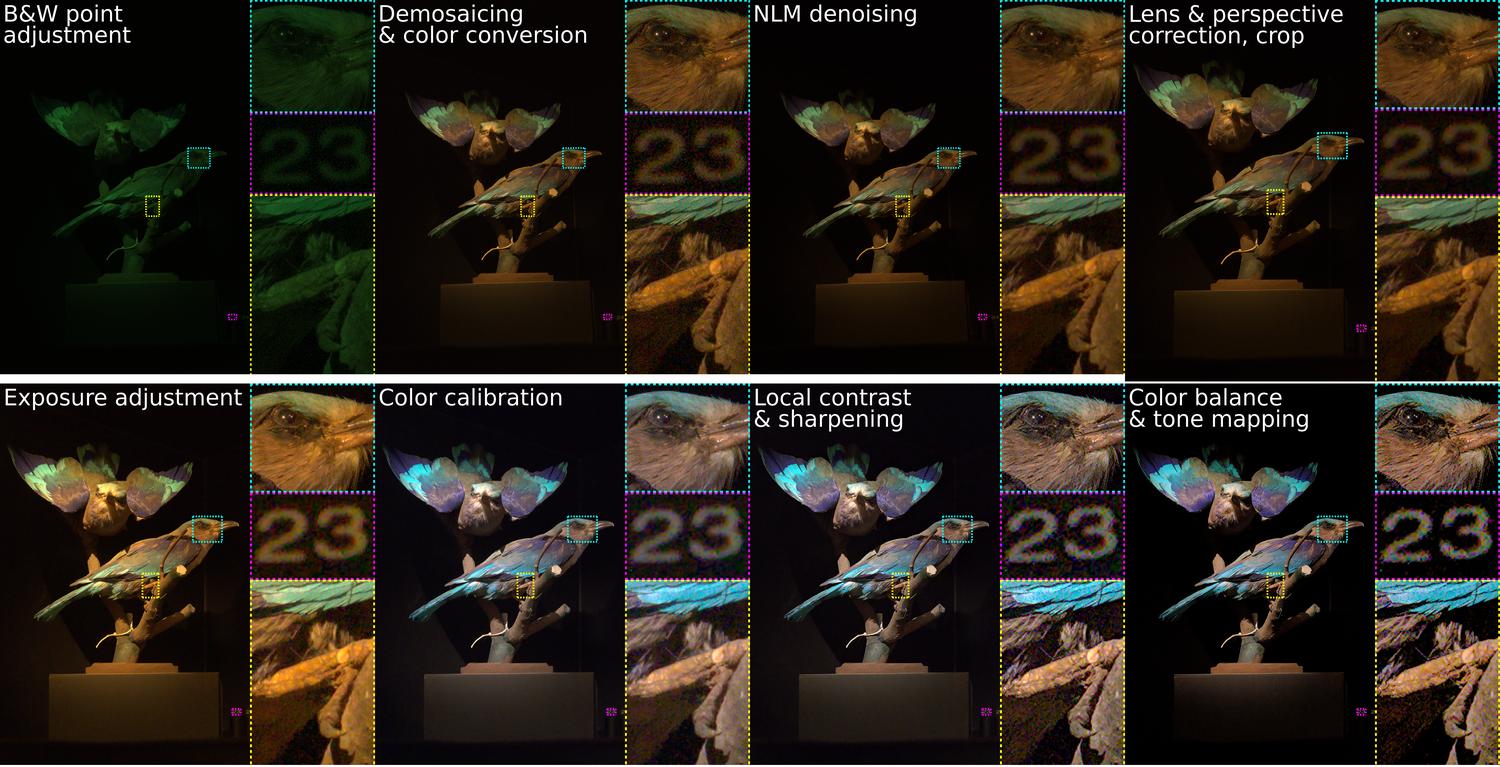}
    \caption[Noisy image development pipeline]{An image development pipeline illustrated using Darktable 5.0 and a moderately noisy image (taken at \acrshort{isocam}16000). \textbf{Throughout this work, we refer to stage 1 as the Bayer input, stage 2 as the Linear \acrshort{rgb} input, and stage 8 as the developed image} (although NLM denoising is shown in step 3 for illustrative purposes, prior denoising is not applied in the development pipeline for the developed images used as input to the models in this work, as shown in \Cref{fig:visualdenoise}.) Darktable was chosen as our primary tool for raw image development due to its wide use within the photography community and its open-source nature, ensuring transparency and reproducible experiments.}
	\label{fig:imgdev}
\end{figure*}

Raw image development converts sensor data into visually interpretable images through steps such as demosaicing, color space transformations, and tone mapping. The variability in raw development pipelines arises from the unique characteristics of camera sensors and user-defined adjustments during development. In-camera \glspl{isp} apply proprietary pipelines optimized for speed and versatility, while raw development software like Darktable (illustrated in \Cref{fig:imgdev}) or Adobe Lightroom offer extensive customization options. These variations pose challenges for denoising models which are often trained on images from a specific raw sensor or development pipeline.

To maintain consistency across sensors, color space transformations are applied to convert camera-specific color spaces (``\acrshort{camrgb}'') to a common working color space such as linear \acrshort{rec2020} \cite{rec2020}. In this study, we hypothesize that models trained on Bayer or linear \acrshort{rgb} data may achieve higher performance by avoiding the non-linearities introduced during image development, and we evaluate this assumption experimentally. Previous studies have not explicitly addressed cross-sensor variability, which we also aim to investigate and address through our proposed methods and dataset. 

\subsection{Learned Image Compression}

Image compression reduces storage and transmission costs while maintaining perceptual quality. Traditional methods like JPEG \cite{jpeg_Wallace_1992}, JPEG2000 \cite{jpeg2000_Skodras_2001}, and JPEG XL \cite{jpegxl_JTC1_2024} rely on transform and entropy coding to exploit redundancies in image data. In these methods, noise increases entropy and reduces compression efficiency.

Learned image compression, pioneered by Ballé et al. \cite{e2e_compression_Balle_2017}, builds upon the \glslink{ae}{autoencoder} architecture \cite{autoencoder_Kramer_1991}, using convolutional \glslink{ae}{autoencoders} and learned entropy models to optimize compression end-to-end. \glslink{ae}{Autoencoders}, initially proposed for dimensionality reduction, form the backbone of learned compression by encoding high-dimensional data into compact latent representations. Subsequent works have improved performance with \acrfull{hp} models, autoregressive priors \cite{autoregression_Minnen_2018}, and transformers \cite{transformer_compression_Lu_2022,transformer_compression_Arezki_2023}. However, these advancements often come with increased computational complexity.

\acrfull{jdc} integrates both tasks into a single framework, as proposed in \cite{compdenoise_Brummer_2023}. By addressing noise during compression, \acrshort{jdc} allocates bits more efficiently to meaningful image content, and improves \acrlong{rd} performance and computational efficiency compared to sequential pipelines. This work extends these principles to raw data, demonstrating the benefits of integrating denoising, demosaicing, and compression into a unified pipeline.


\subsection{Generalization Across Sensors}

Generalizing denoising models across sensors remains a significant challenge due to variations in sensor characteristics, noise patterns, and raw image development pipelines. While datasets like \acrshort{sidd} and \acrshort{nind} have shown that models can generalize across certain cameras when trained on developed images, this is often contingent on consistent development workflows. Dealing with raw data avoids this constraint but introduces inherent variabilities such as sensor-specific noise and color profiles, which require careful handling to ensure robust generalization.

This work explicitly addresses these challenges by introducing a dataset that spans multiple sensors and by developing models capable of generalizing across raw development workflows. We demonstrate the effectiveness of these models in handling diverse sensor characteristics, enabling robust performance across a variety of cameras and workflows. Furthermore, our approach preserves the flexibility inherent to raw data, ensuring compatibility with a wide range of post-processing options.

\section{Proposed Approach}
\label{sec:approach}

This section describes our approach to image denoising and compression, focusing on methods that operate directly on raw Bayer and linear \acrshort{rgb} data. We introduce the \textit{\acrfull{rawnind}}, designed to facilitate research and development on generalized denoising and compression across multiple sensors and workflows. We then detail the proposed methodologies for denoising and compression of raw Bayer data and linear \acrshort{rgb} images, highlighting their advantages over traditional approaches reliant on developed images.

\subsection{RawNIND Dataset}
\label{ssec:proposed_rawnind}

\subsubsection{Dataset Overview}

\begin{figure}[!t]
	\centering
	 \ifthenelse{\boolean{CLASSOPTIONdraftcls}}
	{\includegraphics{figures/histogram-crop.pdf}}
	{
	\includegraphics[width=\columnwidth]{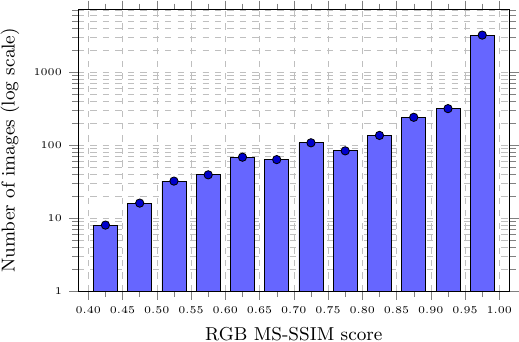}
	\caption[\Acrshort{rawnind} composition by noisiness.]{\acrshort{rawnind} composition by noisiness: histogram of \gls{msssim} between ground-truth and noisy images.}
	\label{fig:histogram}
\end{figure}

\Acrshort{rawnind} comprises paired noisy and clean images captured under diverse conditions using multiple camera sensors. The noisy images were captured at higher \acrshort{isocam} settings and with shorter exposure times compared to their clean counterparts, which were acquired at base \acrshort{isocam} and optimal exposure settings to minimize noise. Each noisy image is paired with one or more corresponding clean image(s) of the same scene, ensuring that both images capture the same content and lighting conditions are consistent. This pairing is achieved by capturing the sequences of noisy and clean images on a sturdy tripod and in rapid succession, minimizing changes in the scene. A preview of the dataset's ground-truth images is shown in \Cref{fig:mosaic}. By retaining the raw sensor data, the dataset preserves linear characteristics and avoids artifacts introduced during image development. To promote generalization across sensors, \acrshort{rawnind} includes data from a variety of Bayer pattern cameras, ensuring compatibility with consumer and professional imaging devices. \Cref{tab:dataset_composition} summarizes the dataset composition, while \Cref{fig:histogram} illustrates the distribution of noise levels. The dataset is published under \url{https://dataverse.uclouvain.be/dataset.xhtml?persistentId=doi:10.14428/DVN/DEQCIM}

\begin{table}[ht]
\centering
\caption{\Acrshort{rawnind} composition by camera model.}
\label{tab:dataset_composition}
\footnotesize
\setlength{\tabcolsep}{2pt}
\conditionalresizebox{%
\begin{tabular}{@{}lrrrr@{}}
\hline
\textbf{Camera} & \textbf{\# Scenes} & \textbf{\# Clean} & \textbf{\# Noisy} & \textbf{\# Images} \\
\hline
Sony ILCE-7C           & 148 & 361  & 1423 & 1784 \\
Canon EOS 500D         & 13  & 13   & 71   & 84   \\
Canon EOS 6D           & 8   & 8    & 80   & 88   \\
Canon EOS 7D           & 7   & 7    & 49   & 56   \\
Canon EOS D60          & 3   & 3    & 12   & 15   \\
Nikon Z 6              & 3   & 3    & 41   & 44   \\
Nikon D40              & 2   & 8    & 8    & 16   \\
Lumix DMC-GH1          & 1   & 1    & 5    & 6    \\
\hdashline
Canon EOS M100         & 11  & 20   & 50   & 70   \\
\hdashline
Fujifilm X-T1 {\tiny (X-Trans)}          & 104 & 128  & 486  & 614  \\
Fujifilm X-T2 {\tiny (X-Trans)}         & 10   & 10    & 54   & 54   \\
\hline
\textbf{Total}         & 310 & 562  & 2279 & 2831 \\
\hline
\end{tabular}
}
\end{table}

\subsubsection{Cross-Sensor Model Consistency}
\begin{figure}[!t]
	\centering
    \ifthenelse{\boolean{CLASSOPTIONdraftcls}}
        {\includegraphics{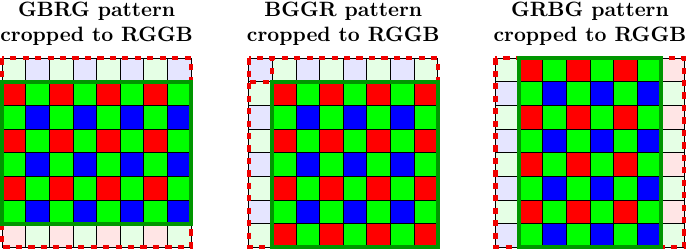}}
        {\includegraphics[width=\columnwidth]{figures/crop_bayer-crop.pdf}}
	
	\caption[Bayer \acrshort{cfa} conversion]{Method for converting various Bayer \acrlong{cfa} patterns (GBRG, BGGR, GRBG) to a consistent RGGB pattern by cropping up to two rows or columns (shown with red dashes). This standardization is crucial for training models that generalize across different camera sensors.}
	\label{fig:crop_bayer}
\end{figure}

Raw images are captured in sensor-specific color spaces (``\acrshort{camrgb}''), which vary across cameras. For models working with linear \acrshort{rgb} images as input, we pre-process the raw data by demosaicing and converting \acrshort{camrgb} to the standardized linear \acrshort{rec2020} color space. Models working with raw Bayer data receive 4-channel Bayer images (\acrshort{camrgb}) as input, preserving the native sensor format. For these models, Bayer images are first cropped to a consistent RGGB pattern using the method illustrated in \Cref{fig:crop_bayer}, ensuring compatibility across cameras with varying Bayer arrangements. The \acrshort{camrgb}-to-\acrshort{rec2020} transformation is applied solely to the model’s 3-channel \acrshort{camrgb} output (upsampled using PixelShuffle) during training before computing the loss function. This ensures consistency with the widely supported \acrshort{rec2020} color space, used as a working profile in software like Darktable. The transformation is based on camera-specific \acrshort{xyz} to \acrshort{camrgb} matrices, calibrated under the D65 (daylight) illuminant and obtained from EXIF metadata, as formalized in \Cref{eq:camrgb2prgb}. This workflow maintains consistency across sensors while preserving flexibility for user-defined workflows.

\begin{subequations}
\begin{align}
\mathbf{M}_{\text{\acrshort{xyz}}\to\text{\glslink{rec2020}{Rec2020}}} &= \begin{bmatrix}
    1.7167 & -0.3557 & -0.2534 \\
    -0.6667 & 1.6165 & 0.0158 \\
    0.0176 & -0.0428 & 0.9422
  \end{bmatrix}
\end{align}
\begin{equation}
  \mathbf{M}_{\text{\glslink{rec2020}{Rec2020}}\to\text{\acrshort{camrgb}}} = \mathbf{M}_{\text{\acrshort{xyz}}\to\text{\glslink{rec2020}{Rec2020}}} \cdot \mathbf{M}_{\text{\acrshort{xyz}}\to\text{\acrshort{camrgb}}}^{-1}
\end{equation}

\begin{equation}
  \text{Image}_{\text{\glslink{rec2020}{Rec2020}}} = \text{Image}_{\text{\acrshort{camrgb}}} \cdot
  \mathbf{M}_{\text{\glslink{rec2020}{Rec2020}}\to\text{\acrshort{camrgb}}}
\end{equation}
\label{eq:camrgb2prgb}
\end{subequations}

\subsection{Proposed Denoising Methods}

We propose two denoising methods tailored to specific representations: raw Bayer images and linear \acrshort{rgb} images.

\subsubsection{Raw Bayer Image Denoising}

This method denoises raw Bayer data directly, deferring demosaicing to the final stage. Denoising is performed on the 4-channel Bayer representation, preserving the native sensor format and significantly reducing computational overhead by operating on one-quarter the spatial dimensions of fully debayered images. The model outputs 3-channel \acrshort{camrgb} data, upsampled to full resolution using PixelShuffle. During training, this output is transformed to linear \acrshort{rec2020} for computing the loss, ensuring compatibility with standard profiles.

\subsubsection{Linear RGB Denoising}

Linear \acrshort{rgb} denoising begins with demosaicing raw images using an edge-aware interpolation method \cite{EA_demosaic_Xin_2005}, or the Markesteijn algorithm for X-Trans images, and converting them to the linear \acrshort{rec2020} color space. This approach avoids non-linearities introduced by different image development modules while maintaining the consistency of a standardized color space, facilitating cross-sensor generalization.

\subsection{Joint Denoising and Compression (with Implicit Demosaicing for Bayer Input)}

To extend the benefits of raw data processing, we integrate denoising and compression into a single framework. This joint approach is investigated with both raw Bayer and linear \acrshort{rgb} image inputs, mirroring the methodology applied in the denoising methods. For raw Bayer input, this inherently involves demosaicing as part of the process, while for linear \acrshort{rgb} input, demosaicing is assumed to have already occurred. Our framework utilizes a convolutional \glslink{ae}{autoencoder} with a learned entropy model \cite{manypriors_Brummer_2021,compdenoise_Brummer_2023} and is trained by minimizing a \acrlong{rd} loss, typically expressed as 
$\mathcal{L} = D(x, \hat{x})+\lambda\text{\gls{bpp}}(\hat{x})$,
where $D$ is a distortion measure between the clean ground-truth image $x$ and the reconstructed image $\hat{x}$ (after \acrlong{jdc} of the input image $y$), $\text{\gls{bpp}}(\hat{x})$ is the bitrate after entropy-coding, and \acrshort{lambda} is a trade-off parameter.  The distortion measure $D$ is typically \gls{mse} or \gls{msssim}, with our work focusing on \gls{msssim}.

For raw Bayer input, the framework integrates \acrfull{jddc}\glsunset{jddc}. By addressing noise at the earliest stage, this joint approach reduces computational overhead and improves \acrlong{rd} performance. Compressing raw Bayer data directly avoids encoding redundant interpolated information, further enhancing compression efficiency. PixelShuffle is used at the end of the decoder stage to reconstruct full-resolution \acrshort{rgb} images from Bayer inputs.

For linear \acrshort{rgb} input, the framework performs \acrfull{jdc} on a demosaiced linear \acrshort{rec2020} input. This approach still benefits from addressing noise early in the pipeline, improving \acrlong{rd} performance compared to compressing noisy linear \acrshort{rgb} images or \acrshort{jdc} of developed images. Furthermore, by operating on linear \acrshort{rgb} data after demosaicing, the framework is agnostic to the specific \acrshort{cfa} pattern of the input sensor, such as Bayer or X-Trans, enhancing its versatility.

\subsection{Advantages of the Proposed Approach}

The proposed methods offer the following advantages:
\begin{itemize}
    \item \textbf{Computational Efficiency:} Raw Bayer denoising and compression reduces computational requirements by operating on one-quarter the spatial dimensions of debayered images.
    \item \textbf{Generalization Across Sensors:} Both Bayer and linear \acrshort{rgb} methods are designed to handle diverse sensor characteristics and raw workflows, ensuring robust performance across devices.
    \item \textbf{Avoidance of Non-Linearities:} By operating on raw or linear \acrshort{rgb} data, our methods avoid artifacts introduced by development modules such as tone mapping, gamma correction, or sharpening. This preserves the intrinsic characteristics of the captured scene and prevents noise amplification, leading to more robust denoising performance.
    \item \textbf{Development Flexibility:} Outputs from Bayer models (3-channel \acrshort{camrgb}) and linear \acrshort{rgb} models (\acrshort{rec2020}) integrate seamlessly into raw development software like Darktable, enabling users to work on denoised images as if they were raw.
    \item \textbf{Improved Compression Efficiency:} The joint framework minimizes redundancy, achieving superior \acrlong{rd} performance compared to traditional methods.
\end{itemize}

\section{Validation Methodology}
\label{sec:experiments}

This section presents the experimental setup, including dataset preparation, baseline models, and evaluation methodologies, to validate the proposed denoising and compression approaches.

\subsection{Dataset Preparation}
\label{subsec:dataset-prep}


\subsubsection{Dataset Preprocessing}

The dataset, which is presented in \Cref{ssec:proposed_rawnind} and downloadable on {\url{https://dataverse.uclouvain.be/dataset.xhtml?persistentId=doi:10.14428/DVN/DEQCIM}}, is stored in two formats: raw Bayer images, processed only to remove empty borders, subtract the black level, and normalize pixel values to the sensor’s white level, and debayered images converted to the linear \acrshort{rec2020} color space. Both formats are divided into overlapping patches of $1024 \times 1024$ pixels ($512 \times 512$ for Bayer) with a stride of $256$ pixels ($128 \times 128$ for Bayer). These patches facilitate efficient loading and flexible filtering during training. Metadata is included with each image pair, including alignment parameters, gain normalization values, alignment loss, and \gls{msssim} scores, to allow flexible dataset filtering based on quality thresholds.

Alignment of noisy and clean images is critical for supervised training. We find the optimal alignment by searching for the translation vector (horizontal and vertical shift) that minimizes the $L1$ loss between gain-normalized images in the \acrshort{rgb} domain (demosaiced images). Aligning in the Bayer domain is impractical due to the single color channel per pixel. The search for the best shift is performed iteratively. Starting from an initial alignment, we explore a small, 3x3 pixel neighborhood of potential shifts around the current best alignment. If a shift within this neighborhood reduces the $L1$ loss, it becomes the new best, and the search continues from that point. This local search is constrained by a maximum search range of $\pm128$ pixels in each direction. The iterative process stops when no better alignment is found within the 3x3 neighborhood or the maximum shift is reached. The resulting optimal shift and its corresponding loss are recorded. Image pairs with an alignment loss exceeding $0.035$ are discarded ($2.4\%$ of the dataset).

Despite careful dataset preparation, issues such as lighting variations, foreign objects or transient elements like insects can occasionally disrupt the image pairs. To further address inconsistencies, binary loss masks are generated to exclude problematic regions from loss computation during training. Masks are applied to pixels with excessive $L1$ loss ($>0.4$ or exceeding the $99.99^{\text{th}}$ percentile of the loss distribution). Overexposed pixels ($\geq 0.99$ in the ground-truth image) are similarly masked. The mask is refined using a binary opening operation to remove small, isolated regions, resulting in smoother and more meaningful masking. Crops with more than $50\%$ of their area masked are excluded.

\subsubsection{Clean data for generalization}
To enhance sensor diversity and improve model generalization, we augmented the dataset with clean (unpaired) raw images from various sources. This augmentation included 561 Bayer images from numerous cameras available on the \texttt{raw.pixls.us} platform, with a maximum \acrshort{isocam} limit of 200 to ensure high image quality. Additionally, we incorporated 11,815 Bayer images captured with several cameras from our personal collection, also restricted to \acrshort{isocam} 200 or lower. These unpaired clean images aim to expose the model to a broader range of sensor characteristics, facilitating improved generalization to unseen sensors. By including these diverse clean images, we evaluate their contribution to training models capable of handling raw data from previously unseen camera sensors.

\subsubsection{Loading train data}
During training, pairs of aligned image patches are loaded. For \acrshort{rgb}-Bayer image pairs, the Bayer shift is halved using integer division. If the shift is odd, an additional row or column is trimmed from both the \acrshort{rgb} and Bayer images to ensure proper alignment. To account for differences in brightness or intensity between the clean and noisy images, their average pixel intensities (gains) are matched. Random $256 \times 256$ ($128 \times 128$ for Bayer) crops are then extracted from the aligned patches.

For models working with developed images, a custom proxy development pipeline is applied to simulate typical image development steps. This pipeline includes logarithmic tone mapping on the luminance channel, edge enhancement using a Laplacian filter, gamma correction, contrast enhancement with a sigmoid function, and image sharpening via Gaussian blur and weighted blending. Parameters and activations for these operations are randomized within plausible ranges, ensuring variability and robustness across a wide range of image characteristics. While this simulated pipeline does not replicate all variations of real-world image development, it provides consistent training data for comparison across methods. The use of this pipeline enables direct evaluation of models trained on raw Bayer, linear \acrshort{rgb}, and developed images representations.

\subsection{Test Sets and Evaluation Conditions}
\label{subsec:test-eval}

\subsubsection{RawNIND test set}
The test set comprises 10 scenes from \acrshort{rawnind}, captured using six different cameras to ensure diverse evaluation conditions. These scenes include \textit{7D-2} (Canon EOS 7D), \textit{Vaxt-i-trad} (Canon EOS 6D), \textit{Pen-pile} (Panasonic DMC-GH1), \textit{MuseeL-vases-A7C}, \textit{TitusToys}, \textit{boardgames\_top}, \textit{Laura\_Lemons\_platformer}, and \textit{MuseeL-bluebirds-A7C} (all captured with the Sony A7C), as well as \textit{D60-1} (Canon EOS D60) and \textit{MuseeL-Saint-Pierre-C500D} (Canon EOS 500D). This diverse selection of scenes and camera models ensures representation of varying sensor characteristics and noise profiles.

To evaluate generalization across entirely unseen sensors, an additional test set includes six scenes captured with the Canon EOS M100 that has not been seen at training, and referred to as the ``unknown sensor'' test set. These scenes are denoted \textit{LucieB\_bw\_drawing1}, \textit{LucieB\_bw\_drawing2}, \textit{LucieB\_board}, \textit{LucieB\_painted\_wallpaper}, \textit{LucieB\_painted\_plants}, and \textit{LucieB\_groceries}.

\subsubsection{Evaluation pipeline}
Raw model outputs are not evaluated directly due to their lack of immediate visual usability. Instead, outputs are developed using Darktable with manually curated settings. The same development instructions are applied to raw ground-truth images and model outputs, enabling consistent comparisons using the \gls{msssim} metric. For models trained to take developed images as input, the same development steps are applied directly to their input images. This evaluation method may slightly favor models processing developed images, as their outputs are assessed directly. When compressing clean images, these models only need to reconstruct the input, whereas the outputs of models handling raw or linear \acrshort{rgb} data undergo development where imperfections can be amplified differently. Despite this, applying the same development process to all outputs reflects real-world workflows and allows for the fairest possible comparison.

The sidecar \acrshort{xmp} files containing the manually curated settings for test images are published alongside the dataset, allowing reproducibility of results and enabling further analysis or experimentation by the community.

\subsection{Baseline Models for Denoising}
\label{subsec:baseline-denoising}

We evaluate denoising models trained on raw Bayer images, linear \acrshort{rgb} images, and developed images (``\acrshort{srgb}''). The base model for developed and linear RGB images is a variant of the U-Net architecture, with the number of channels halved, reducing computational complexity by a factor of four. \gls{relu} activations are replaced with Leaky \gls{relu} (with a negative slope of $0.2$).

For Bayer images, which operate on one-quarter of the pixels, a PixelShuffle upscaling layer \cite{pixelshuffle_Wenzhe_2016} is added at the end of the network, resulting in an additional fourfold reduction in complexity. Unless otherwise specified, all U-Net models in this work refer to this simplified variant. Additional variants in the ablation study include models with:
\begin{itemize}
    \item \textbf{Extra Pairs:} Incorporating additional noisy-clean pairs from the unknown sensor.
    \item \textbf{No Clean Data:} Excluding unpaired clean images during training (i.e. denoising-only).
    \item \textbf{Pre-Upsampled:} Bilinear upsampling of Bayer data before input.
    \item \textbf{Gamma Correction:} Applying the gamma correction \(\text{output} \leftarrow \text{output}^{1 / 2.2}\) for \(\text{output} > 0\) before loss computation.
    \item \textbf{More channels:} Expanding channels by $1.5\times$ over the base model.
\end{itemize}
\gls{bm3d} is used as a traditional baseline for linear \acrshort{rgb} and fully developed inputs.

\subsection{Joint Denoising and Compression Experiments}
\label{subsec:exp-compression}

We evaluate \acrshort{jdc} models trained on Bayer, linear \acrshort{rgb}, and developed images. The base model for fully developed and linear \acrshort{rgb} images is the compression \glslink{ae}{autoencoder} architecture defined in \cite{manypriors_Brummer_2021} and used in for standard \acrshort{jdc} in \cite{compdenoise_Brummer_2023}, while the Bayer models get one additional convolution and a PixelShuffle layer. Variants include pre-upsampling Bayer inputs and excluding clean data. Compression-only models and sequential pipelines (denoise first using the base U-Net from \Cref{subsec:baseline-denoising}, then compress) are also evaluated. The computational complexity of compression encoders as well as that of denoising models is described in \Cref{tab:model_complexity}. The complexity of the compression decoder is not significantly reduced because, despite working with reduced Bayer dimensions, the last layer involves a final convolution which takes place in high-resolution; further experiments are needed to reduce the complexity of the compression decoder. For comparison, JPEG XL is tested on linear \acrshort{rgb} and developed images, with \acrlong{rd} curves used to assess performance across noise levels.

\begin{table}
    \centering
    \caption{Model complexity for denoising and compression}
    \label{tab:model_complexity}
    \conditionalresizebox{%
        \begin{tabular}{l l r}
            \textbf{Model} & \textbf{Input Image Type} & \textbf{\acrshort{gmac}/\acrshort{mp}} \\
            \midrule
            \multicolumn{3}{l}{\textbf{Denoising}} \\
            \rowcolor{matplotlibgreen!10} U-Net (½ \# of channels)\textsuperscript{(1)} & Bayer & 55 \\
            U-Net (½ \# of channels)\textsuperscript{(2)} & \acrshort{rgb} (lin. or dev.) & 220 \\
            U-Net (all channels)\textsuperscript{(3)} \cite{unet_Ronneberger_2015} & \acrshort{rgb} (lin. or dev. \cite{nind_Brummer_2019}) & 875 \\
            \midrule
            \multicolumn{3}{l}{\textbf{Compression encoder}} \\
            \rowcolor{matplotlibgreen!10} \acrshort{jddc} & Bayer & 22 \\
            \acrshort{jddc} (pre-upsampled) & Bayer & 87 \\
            \acrshort{jdc} & \acrshort{rgb} (lin. or dev. \cite{compdenoise_Brummer_2023}) & 86 \\
            Denoise\textsuperscript{(1)} then compress & Bayer & 141 \\
            Denoise\textsuperscript{(2)} then compress & \acrshort{rgb} (lin. or dev.) & 305 \\
            Denoise\textsuperscript{(3)} then compress & \acrshort{rgb} (lin. or dev.) & 1095 \\
        \end{tabular}
    }
\end{table}

Results are evaluated using developed outputs to ensure fair comparisons across all formats. The \gls{msssim} and \gls{bpp} metrics capture image quality and compression efficiency. \gls{rd} curves are generated for the \acrshort{rawnind} test set, highlighting the impact of noise, (non-linear) image development, and the advantages of integrating denoising into the compression pipeline.

\section{Results and Discussion}
\label{sec:results}

\subsection{Denoising Performance}

\begin{figure*}[t!]
	\centering
    \ifthenelse{\boolean{CLASSOPTIONdraftcls}}
        {\includegraphics[width=.77\textwidth]{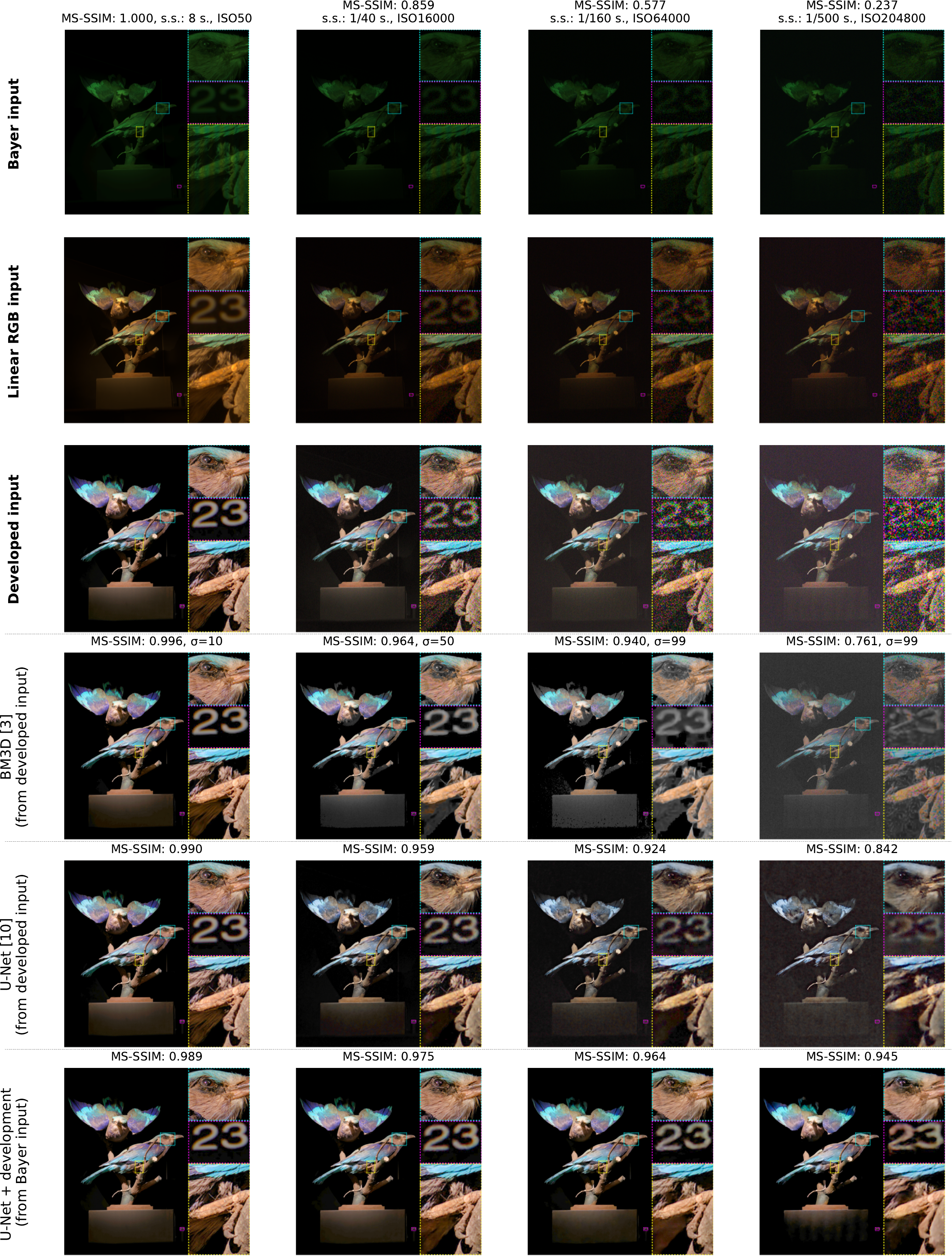}}
        {\includegraphics[width=.96\textwidth]{figures/Picture1_32.yaml_denoising_jddc-crop.pdf}}
	\caption{Visual and \gls{msssim} results for denoising a test image from the \acrshort{rawnind} main test set (\texttt{MuseeL-bluebirds-A7C}) under varying input noise levels. Results are shown for different denoising methods and input formats. Development is performed using darktable-cli and the same set of development instructions (\acrshort{xmp} file) whose steps are illustrated in \Cref{fig:imgdev}.}
	\label{fig:visualdenoise}
\end{figure*}

The proposed denoising methods were evaluated using raw Bayer, linear \acrshort{rgb}, and developed images representations. \Cref{tab:denoising} presents the \gls{msssim} indices across the \acrshort{rawnind} test set and an additional ``unknown sensor'' test set. \ifthenelse{\boolean{arxivflag}}{\Cref{fig:visualdenoise,fig:visualdenoise_unk}}{\Cref{fig:visualdenoise}} show quantitative and qualitative results on two test images.

\ifthenelse{\boolean{CLASSOPTIONdraftcls}}{

\begin{table*}[!t] 
    \centering
    \captionsetup[subtable]{justification=centering}
    \setlength{\tabcolsep}{3pt} 

    \subfloat[\small{Main Models}]{
        \begin{tabular}{l *{3}{S[table-format=1.3]} H *{2}{S[table-format=1.3]} | c}
            & \multicolumn{6}{c|}{\textbf{\acrshort{rawnind} test set}} & {\textbf{Unknown sensor}} \\
            \textbf{Denoising method (input image)} & {0.468} & {0.598} & {0.680} & {0.763} & {0.846} & {0.933} & {0.686} \\
            \midrule
            U-Net (Bayer, ours) & \bfseries 0.930 & \bfseries 0.941 & \bfseries 0.949 & \bfseries 0.956 & \bfseries 0.962 & \bfseries 0.972 & \bfseries 0.873 \\
            U-Net (linear \acrshort{rgb}, ours) & \bfseries 0.930 & \bfseries 0.942 & \bfseries 0.949 & \bfseries 0.957 & \bfseries 0.963 & \bfseries 0.973 & \bfseries 0.874 \\
            U-Net (developed images, \cite{compdenoise_Brummer_2023}) & 0.862 & 0.891 & 0.910 & 0.928 & 0.945 & 0.967 & 0.836 \\
            \acrshort{bm3d} \cite{bm3d_Dabov_2006} (linear \acrshort{rgb}) & 0.838 & 0.873 & 0.893 & 0.906 & 0.921 & 0.940 & - \\
            \acrshort{bm3d} \cite{bm3d_Dabov_2006} (developed images) & 0.826 & 0.870 & 0.895 & 0.919 & 0.944 & 0.968 & 0.823 \\
        \end{tabular}
    }

    \subfloat[\small{Ablation Study}]{
        \begin{tabular}{l *{3}{S[table-format=1.3]} H *{2}{S[table-format=1.3]} | c}
            & \multicolumn{6}{c|}{\textbf{\acrshort{rawnind} test set}} & {\textbf{Unknown sensor}} \\
            \textbf{Input image (training method)} & {0.468} & {0.598} & {0.680} & {0.763} & {0.846} & {0.933} & {0.686} \\
            \midrule
            Bayer (extra pairs) & 0.925 & 0.938 & \bfseries 0.946 & \bfseries 0.954 & \bfseries 0.961 & \bfseries 0.972 & \bfseries 0.876 \\
            Bayer (more channels) & \bfseries 0.929 & \bfseries 0.940 & \bfseries 0.947 & \bfseries 0.954 & \bfseries 0.960 & 0.969 & - \\
            Bayer (no clean data) & 0.926 & \bfseries 0.938 & \bfseries 0.946 & \bfseries 0.954 & \bfseries 0.961 & \bfseries 0.971 & 0.871 \\
            Bayer (pre-upsampled) & \bfseries 0.931 & \bfseries 0.943 & \bfseries 0.950 & \bfseries 0.958 & \bfseries 0.964 & \bfseries 0.973 & \bfseries 0.877 \\
            Bayer (with gamma) & 0.910 & 0.928 & 0.939 & 0.950 & \bfseries 0.961 & \bfseries 0.973 & 0.871 \\
            Linear \acrshort{rgb} (extra pairs) & \bfseries 0.929 & \bfseries 0.940 & \bfseries 0.947 & \bfseries 0.955 & \bfseries 0.960 & \bfseries 0.970 & \bfseries 0.873 \\
            Linear \acrshort{rgb} (no clean data) & \bfseries 0.929 & \bfseries 0.940 & \bfseries 0.947 & \bfseries 0.955 & \bfseries 0.960 & \bfseries 0.970 & 0.870 \\
            Linear \acrshort{rgb} (with gamma) & \bfseries 0.931 & \bfseries 0.942 & \bfseries 0.950 & \bfseries 0.958 & \bfseries 0.963 & \bfseries 0.973 & \bfseries 0.875 \\
        \end{tabular}
    }

    \caption{Average MS-SSIM scores of denoised images from the RawNIND main and unknwon sensor test sets, after manual development with Darktable. Columns correspond to subsets with varying noise levels (average MS-SSIM of developed input images), with the highlighted column representing the entire main test set. Bold values are within 99.5\% of the best score in each column. All models use the U-Net architecture, except BM3D. The second sub-table presents an ablation study exploring different configurations of the Bayer and Linear \acrshort{rgb} models. The models configurations used in this table are described in \Cref{subsec:baseline-denoising}.}\label{tab:denoising}
\end{table*}
}{
\begin{table*}[ht]
    \centering
    \captionsetup[subtable]{justification=centering}
    
    \subfloat[\small{Main Models}]{
        \begin{tabular}{lcccHcc|c}
            & \multicolumn{6}{c|}{\textbf{\acrshort{rawnind} test set}} & \textbf{Unknown sensor} \\
            \textbf{Denoising method (input image)} & 0.468 & 0.598 & 0.680 & 0.763 & 0.846 & 0.933 & 0.686 \\
            \midrule
            U-Net (Bayer, ours) & \textbf{0.930} & \textbf{0.941} & \textbf{0.949} & \textbf{0.956} & \textbf{0.962} & \textbf{0.972} & \textbf{0.873} \\
            U-Net (linear \acrshort{rgb}, ours) & \textbf{0.930} & \textbf{0.942} & \textbf{0.949} & \textbf{0.957} & \textbf{0.963} & \textbf{0.973} & \textbf{0.874} \\
            U-Net (developed images, \cite{compdenoise_Brummer_2023}) & 0.862 & 0.891 & 0.910 & 0.928 & 0.945 & 0.967 & 0.836 \\
            \acrshort{bm3d} \cite{bm3d_Dabov_2006} (linear \acrshort{rgb}) & 0.838 & 0.873 & 0.893 & 0.906 & 0.921 & 0.940 & - \\
            \acrshort{bm3d} \cite{bm3d_Dabov_2006} (developed images) & 0.826 & 0.870 & 0.895 & 0.919 & 0.944 & 0.968 & 0.823 \\
        \end{tabular}
    }
    
    \hfill
    
    \subfloat[\small{Ablation Study}]{
        \begin{tabular}{lcccHcc|c}
            & \multicolumn{6}{c|}{\textbf{\acrshort{rawnind} test set}} & \textbf{Unknown sensor} \\
            \textbf{Input image (training method)} & 0.468 & 0.598 & 0.680 & 0.763 & 0.846 & 0.933 & 0.686 \\
            \midrule
            Bayer (extra pairs) & 0.925 & 0.938 & \textbf{0.946} & \textbf{0.954} & \textbf{0.961} & \textbf{0.972} & \textbf{0.876} \\
            Bayer (more channels) & \textbf{0.929} & \textbf{0.940} & \textbf{0.947} & \textbf{0.954} & \textbf{0.960} & 0.969 & - \\
            Bayer (no clean data) & 0.926 & \textbf{0.938} & \textbf{0.946} & \textbf{0.954} & \textbf{0.961} & \textbf{0.971} & 0.871 \\
            Bayer (pre-upsampled) & \textbf{0.931} & \textbf{0.943} & \textbf{0.950} & \textbf{0.958} & \textbf{0.964} & \textbf{0.973} & \textbf{0.877} \\
            Bayer (with gamma) & 0.910 & 0.928 & 0.939 & 0.950 & \textbf{0.961} & \textbf{0.973} & 0.871 \\
            Linear \acrshort{rgb} (extra pairs) & \textbf{0.929} & \textbf{0.940} & \textbf{0.947} & \textbf{0.955} & \textbf{0.960} & \textbf{0.970} & \textbf{0.873} \\
            Linear \acrshort{rgb} (no clean data) & \textbf{0.929} & \textbf{0.940} & \textbf{0.947} & \textbf{0.955} & \textbf{0.960} & \textbf{0.970} & 0.870 \\
            Linear \acrshort{rgb} (with gamma) & \textbf{0.931} & \textbf{0.942} & \textbf{0.950} & \textbf{0.958} & \textbf{0.963} & \textbf{0.973} & \textbf{0.875} \\
        \end{tabular}
    }
    \caption{Average MS-SSIM scores of denoised images from the RawNIND main and unknwon sensor test sets, after manual development with Darktable. Columns correspond to subsets with varying noise levels (average MS-SSIM of developed input images), with the highlighted column representing the entire main test set. Bold values are within 99.5\% of the best score in each column. All models use the U-Net architecture, except BM3D. The second sub-table presents an ablation study exploring different configurations of the Bayer and Linear \acrshort{rgb} models. The models configurations used in this table are described in \Cref{subsec:baseline-denoising}.}
    \label{tab:denoising}
\end{table*}
}


\ifthenelse{\boolean{arxivflag}}
{
\begin{figure*}[hbtp]
	\centering
	\includegraphics[width=1.\textwidth]{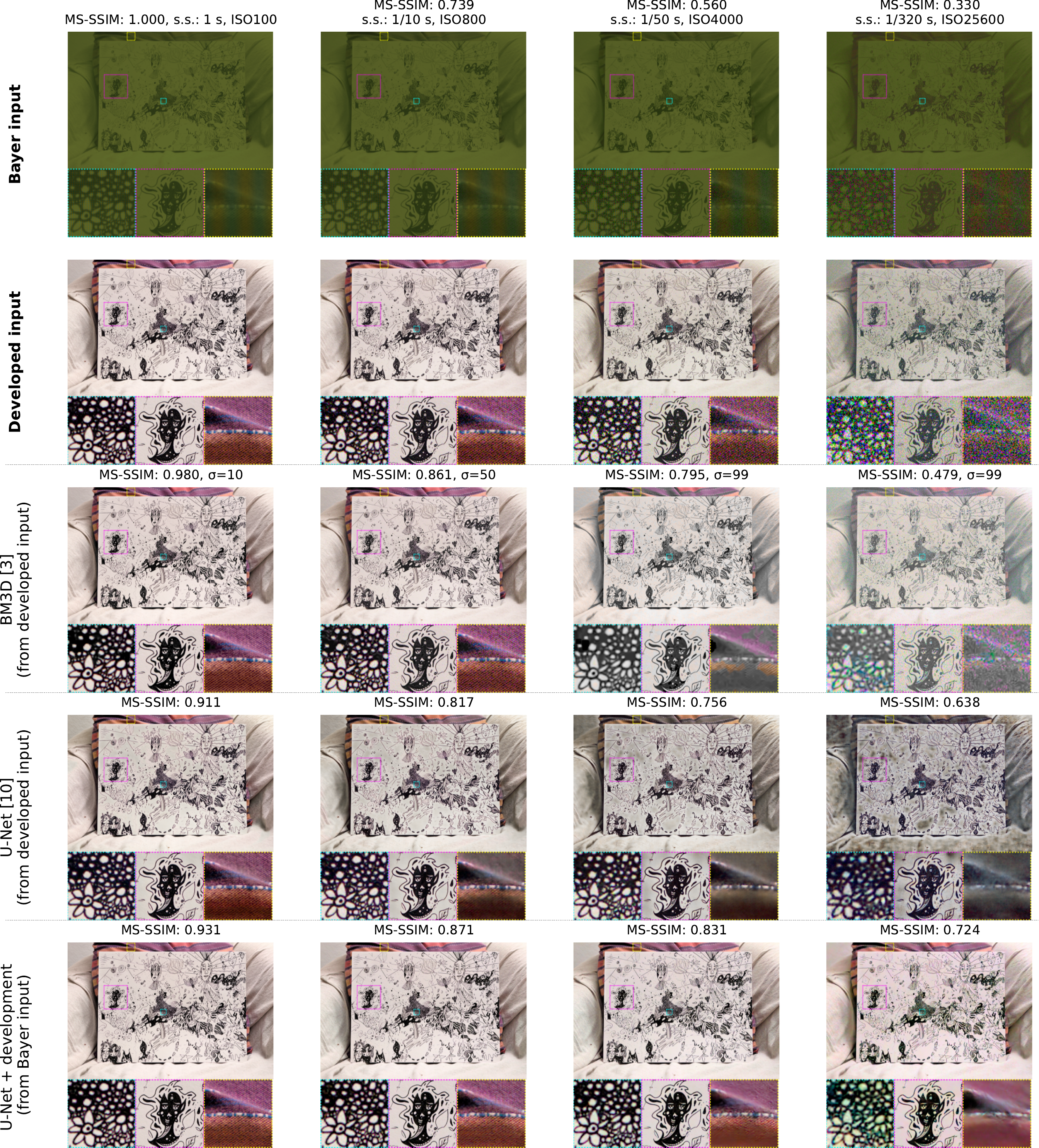}
	\caption[Denoising an unknown sensor test image]{Visual and \gls{msssim} results for denoising an image from the \acrshort{rawnind} unknown sensor test set (\texttt{LucieB\_bw\_drawing1}) under varying input noise levels. Development is performed using darktable-cli and the same set of development instructions (\acrshort{xmp} file).}
	\label{fig:visualdenoise_unk}
\end{figure*}
}
{}

\textbf{Advantages of Raw and Linear RGB Data:}  
Models denoising raw Bayer and linear \acrshort{rgb} images consistently outperformed those trained on developed (``\acrshort{srgb}'') images in \gls{msssim}. This underscores the advantage of working with raw or linear \acrshort{rgb} data, as it retains more linear characteristics and avoids the artifacts introduced in image development.

\textbf{Bayer vs. Linear RGB:}
The performance of Bayer and linear \acrshort{rgb} models was comparable, with Bayer models achieving a significant edge in computational efficiency by processing four-channel Bayer data at reduced spatial dimensions. This approach reduces the computational complexity by a factor of 4, as Bayer models require only a quarter of the computations needed for fully debayered linear \acrshort{rgb} images. The use of PixelShuffle to upscale Bayer outputs ensures high-quality results without adding significant computational overhead.

\textbf{Generalization to Unknown Sensors:}  
The models demonstrated strong cross-sensor performance. Bayer models trained with ``extra pairs'' from the unknown sensor achieved an average \gls{msssim} of 0.876 on the unknown sensor test set, while just adding clean images from different sensors resulted in an \gls{msssim} of 0.873 and training without clean data yield a score of 0.871. This marginal difference highlights that incorporating clean data aids the Bayer model's generalization, but the improvement is minimal. Notably, Bayer models generalize well when trained exclusively with \acrshort{rawnind}. For Linear \acrshort{rgb} models, the inclusion of clean training data provided no benefit.

\textbf{Gamma Correction:}  
Applying gamma correction before loss computation did not yield significant benefits and, in some cases, harmed training stability. Bayer models experienced a 0.6\% \gls{msssim} drop, likely due to the mismatch between the corrected outputs and the intrinsic characteristics of raw data.

\subsection{Compression Performance}

We assessed the compression performance by evaluating the \acrlong{rd} curves of our models. \Cref{fig:rd_manproc_combined} shows the \gls{bpp} versus \gls{msssim} performance for different models on the \acrshort{rawnind} test set under various noise conditions.

\begin{figure*}
	\centering
	\ifthenelse{\boolean{CLASSOPTIONdraftcls}}
        {\includegraphics[width=.86\textwidth]{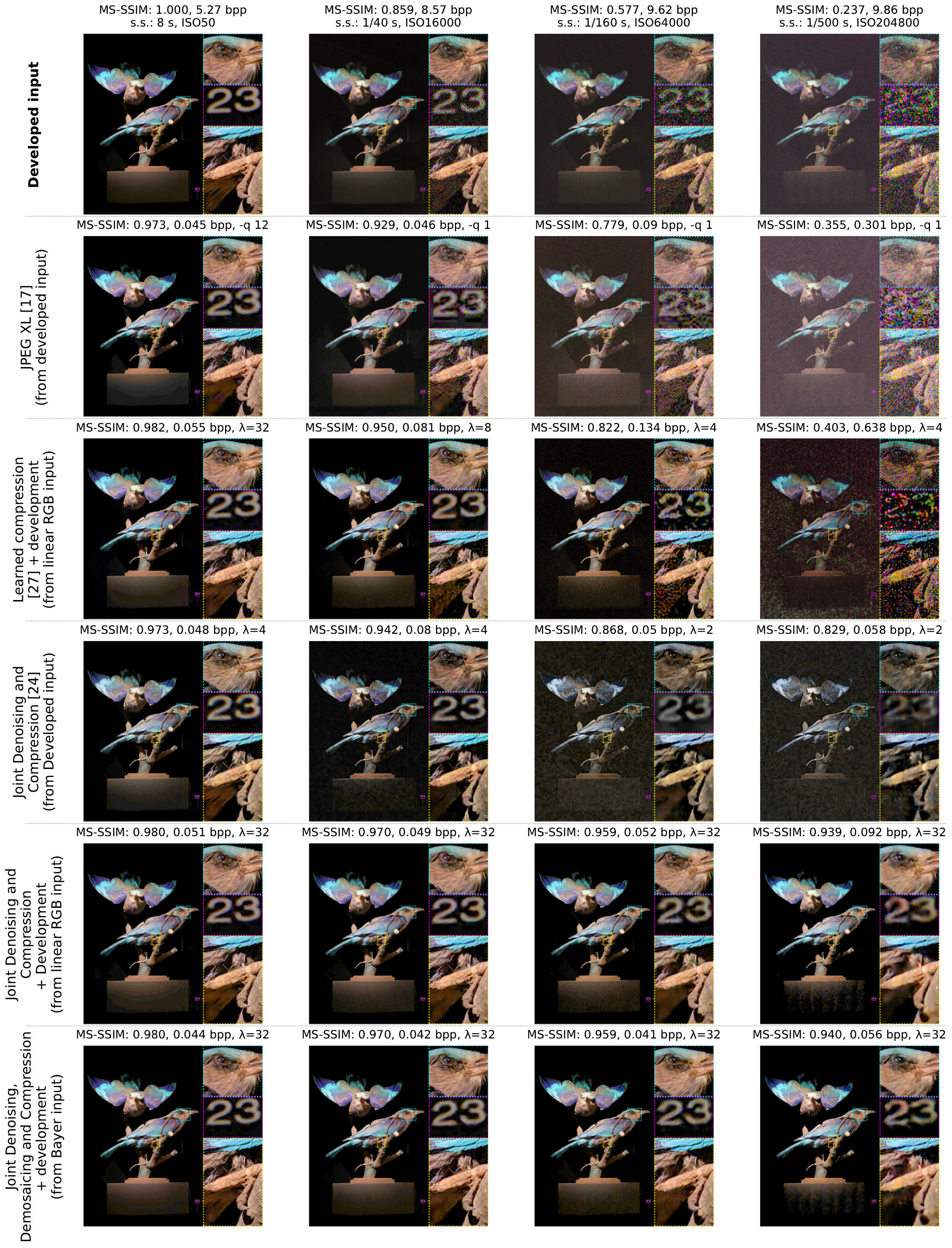}}
        {\includegraphics[width=.97\textwidth]{figures/Picture1_32.yaml_compression_jddc.pdf}}
	\caption[Compression, \acrshort{jdc}, and \acrshort{jddc} of a test image]{\Acrlong{rd} and visual results for compressing a test image (\texttt{MuseeL-bluebirds-A7C}) under varying input noise levels. Results are presented for different compression methods and input formats. The input \gls{bpp} is measured using lossless xz compression on the raw image. See \Cref{fig:visualdenoise} for all input formats.}
	\label{fig:visualcompress}
\end{figure*}

\newlength{\rdmanprocfigurewidth}
\ifthenelse{\boolean{CLASSOPTIONdraftcls}}
    {\setlength{\rdmanprocfigurewidth}{.46\textwidth}}  
    {\setlength{\rdmanprocfigurewidth}{.95\columnwidth}}  
\begin{figure}[!t]
    \centering
    \begin{tabular}{c}
        \raisebox{0.9\height}{\hspace{-10pt}(a)}%
        \label{fig:rd_manproc_a}
        \includegraphics[width=\rdmanprocfigurewidth]{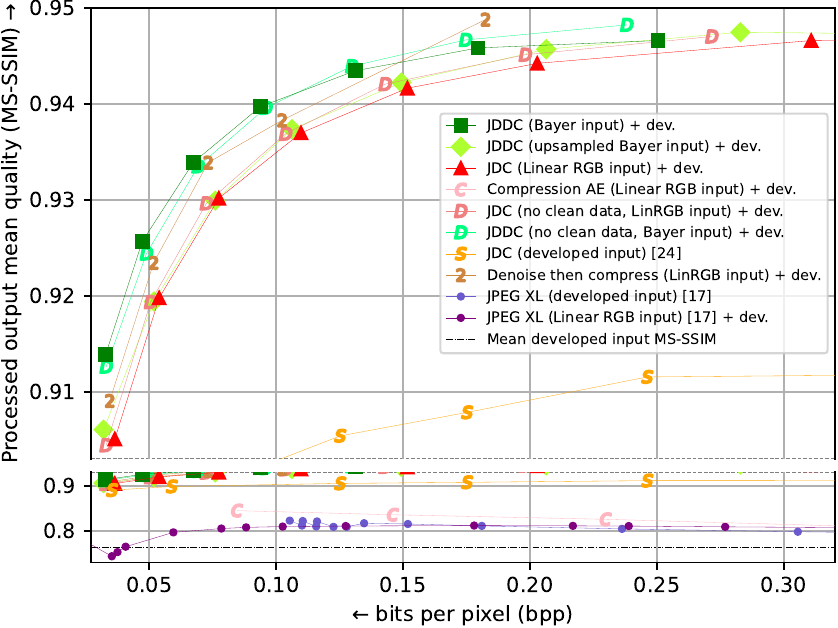} \\
        \raisebox{0.9\height}{\hspace{-10pt}(b)}%
        \label{fig:rd_manproc_b}
        \includegraphics[width=\rdmanprocfigurewidth]{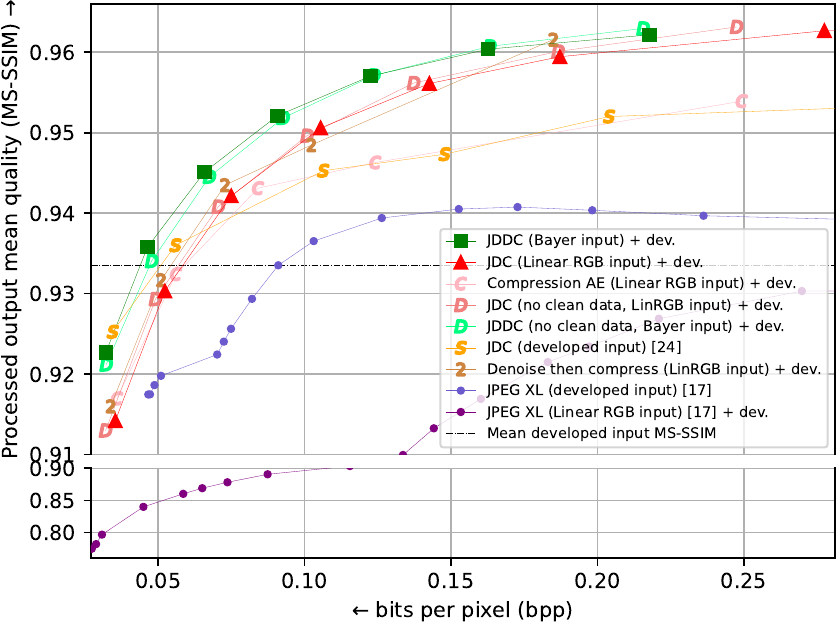} \\
        \raisebox{0.9\height}{\hspace{-10pt}(c)}%
        \label{fig:rd_manproc_c}
        \includegraphics[width=\rdmanprocfigurewidth]{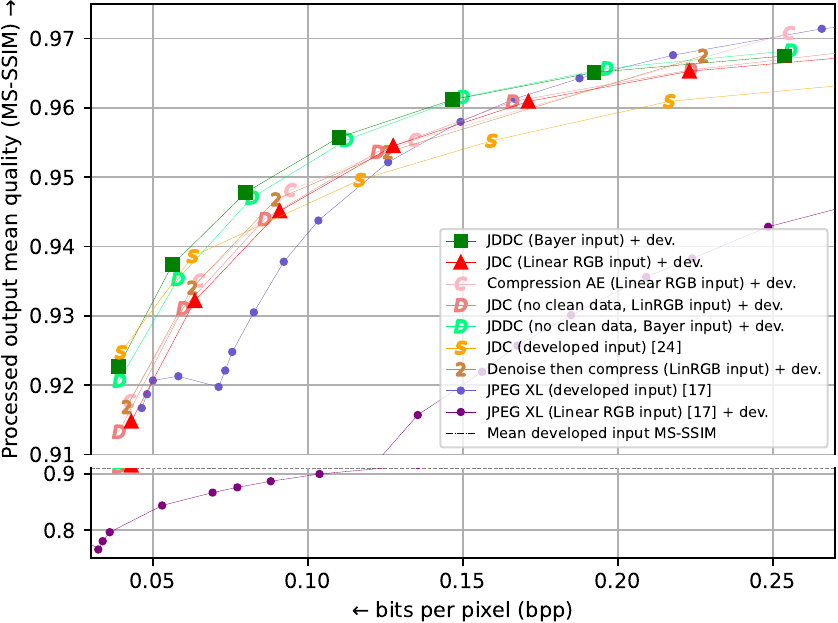} \\
    \end{tabular}
    \caption[Rate-distortion of different methods on the \acrshort{rawnind} test set with different input formats and noise levels.]{\Acrlong{rd} (\gls{bpp} vs. \gls{msssim}) performance of different models on the \acrshort{rawnind} test set after manual development with Darktable. (a) Entire \acrshort{rawnind} test set (163 image pairs, $\overline{\mathrm{\text{\gls{msssim}}}}=0.763$). (b) Low-noise images (80 image pairs, $\overline{\mathrm{\text{\gls{msssim}}}}=0.933$). (c) Ground-truth (clean) images only (15 images, $\overline{\mathrm{\text{\gls{msssim}}}}=1.00$).}
    \label{fig:rd_manproc_combined}
\end{figure}

Our results show that models trained to compress Bayer data consistently achieve superior \acrlong{rd} performance compared to those compressing linear \acrshort{rgb} or developed (``\acrshort{srgb}'') images, regardless of noise level. Notably, \acrlong{jdc} of developed images perform significantly worse than models using linear \acrshort{rgb} or Bayer data, underscoring the limitations of compressing developed images in the presence of noise.

As shown in \Cref{fig:rd_manproc_combined}(a), the \acrlong{jddc} model generally outperforms sequential models that apply denoising followed by compression. However, consistent with the findings of \cite{compdenoise_Brummer_2023}, the sequential approach achieves better \acrlong{rd} performance at the highest bitrates in the presence of strong noise, suggesting that U-Net is a more effective denoising architecture. Merely increasing the number of channels did not yield any improvements in the \acrshort{jddc} model.

Compression-only models, including those compressing linear \acrshort{rgb} images or using standard codecs like JPEG XL, struggle significantly in the presence of noise. Image quality even degrades as bitrate increases, as these methods reconstruct more of the input noise. Interestingly, these compression-only models perform some implicit denoising; although this denoising is suboptimal, the compressed noisy images often have slightly higher image quality than the original noisy inputs. However, this implicit denoising is insufficient for practical purposes and does not replace the need for explicit denoising. In our experiments, we were unable to reliably train a Bayer compression-only model, indicating that integrating denoising helps stabilize training and improve results in the raw domain. This underscores the necessity of denoising to achieve effective compression, particularly for raw Bayer data.

Additionally, models compressing upscaled Bayer images (using bilinear interpolation) achieve \acrlong{rd} performance that is slightly better than models receiving debayered linear \acrshort{rgb} inputs, which use more complex edge-aware interpolation methods. The performance of these upscaled Bayer models falls between that of the raw Bayer models and the linear \acrshort{rgb} models.

Compressing linear \acrshort{rgb} images with JPEG XL and then developing them yields worse results than compressing developed images directly. This aligns with expectations, as JPEG XL is optimized for developed images, and applying it to linear \acrshort{rgb} data without adjustments results in suboptimal performance. These findings emphasize the importance of aligning the compression method with the characteristics of the image data.

For clean images (\Cref{fig:rd_manproc_combined}(c)), the performance gap between the models narrows. Both Bayer and linear \acrshort{rgb} joint models perform well, even when trained exclusively on noisy datasets without clean images. Notably, Bayer models still outperform linear \acrshort{rgb} models, even those trained solely for image compression. ``\acrshort{srgb}'' models exhibit a slight advantage over linear \acrshort{rgb} models at the lowest bitrates.

Compression-only methods perform adequately on clean images but suffer when even small amounts of noise are present, as shown in \Cref{fig:rd_manproc_combined}(b). This reinforces the importance of incorporating denoising in compression pipelines for robust performance across varying noise levels.

We also tested training compression models with gamma correction applied before the loss function, but this approach consistently resulted in poor performance, with models ceasing to improve early in training. This suggests that applying gamma correction is not only ineffective but also detrimental to the training process.

\subsection{Insights and Implications}

Our experiments demonstrate the advantages of processing raw images over developed images for both denoising and compression tasks. Models operating on raw Bayer data or linear \acrshort{rgb} images consistently outperform those trained on developed images, particularly in the presence of noise.

The comparable performance between Bayer and linear \acrshort{rgb} models indicates that both methods are effective for denoising. However, Bayer models offer substantial computational savings in both denoising and compression by operating on data with reduced spatial dimensions, processing only one-quarter of the pixels compared to debayered \acrshort{rgb} images.

Incorporating clean images from diverse sensors slightly enhances generalization to unknown sensors, especially for Bayer models. The minimal performance drop when working with images from unseen sensors suggests that the models inherently generalize well across different camera types, reducing the need for extensive data acquisition and retraining across devices.

In compression tasks, integrating denoising is essential for effective performance, as compression-only models struggle with noisy data. We've also observed increasing stability in training models with \acrlong{jdc} versus those trained for compression only.
In addition to being less computationally expensive, Bayer compression models offer superior \acrlong{rd} performance as they operate directly on meaningful image data without the overhead introduced when encoding interpolated \acrshort{rgb} values. This hypothesis is supported by the performance of the "pre-upsampled" bilinear Bayer model, which falls between that of Bayer and linear \acrshort{rgb} models; simpler upsampling appears to make it easier for the model to discard interpolated data.

For practical applications, these findings suggest that revisiting traditional image development pipelines to incorporate denoising and compression at the raw data level can lead to significant improvements in image quality and data efficiency. One potential workflow involves compressing raw images and transmitting \glslink{xmp}{sidecar} files containing development instructions, allowing for non-destructive editing while leveraging the benefits of raw data compression. Furthermore, integrating denoising models directly into the raw image development pipeline ensures compatibility with any image development workflow.

\section{Conclusion}
\label{sec:conclusion}

This work addresses the limitations of denoising and compression on developed images by introducing the \acrfull{rawnind} and demonstrating the clear advantages of operating directly on raw or linear data. We consistently showed that models using raw Bayer and linear \acrshort{rgb} data outperform those using developed images, highlighting the benefits of preserving linearity and avoiding development artifacts. \acrshort{rawnind} enabled the development of denoising models that effectively generalize across diverse camera sensors and image development workflows. Furthermore, our exploration of \acrlong{jdc} at the raw data level showcases substantial improvements in \acrlong{rd} performance and computational efficiency. This raw-centric approach not only enhances image quality and compression efficiency but also supports flexible, non-destructive editing workflows through the storage of compressed raw data with accompanying development metadata. The comparable performance between Bayer and linear \acrshort{rgb} models indicates that both methods are effective for denoising. However, Bayer models offer significant computational savings in both denoising and compression by operating on data with reduced spatial dimensions, processing only one-quarter of the pixels compared to debayered \acrshort{rgb} images. Incorporating clean images from diverse sensors slightly enhances generalization to unknown sensors, especially for Bayer models. In compression tasks, integrating denoising is essential for effective performance, as compression-only models struggle with noisy data, and our joint models show increased training stability. Bayer compression models offer superior \acrlong{rd} performance by operating directly on meaningful image data without interpolative overhead. Future research should explore further architectural optimizations for raw data processing and the integration of these techniques into practical image processing systems. To foster further innovation and collaboration, \acrshort{rawnind} and its associated tools are publicly available.

\ifthenelse{\boolean{CLASSOPTIONdraftcls}}
{}
{
\section{Acknowledgements}
This research has been funded by the Walloon Region and intoPIX. C. De Vleeschouwer is funded by the `Fond de la Recherche Scientifique de Belgique (F.R.S.-FNRS)'. Computational resources have been provided by the supercomputing facilities of the Université catholique de Louvain (CISM/UCL) and the Consortium des Équipements de Calcul Intensif en Fédération Wallonie Bruxelles (CÉCI) funded by the Fond de la Recherche Scientifique de Belgique (F.R.S.-FNRS) under convention 2.5020.11 and by the Walloon Region. 31 of the 310 \acrshort{rawnind} scenes were captured by the following contributors: Iain Fergusson, Matti Blume, Peter Wemmert, and ``sillyxone''. We thank Johannes Hanika for his insights on color conversions.
}
 
\bibliographystyle{IEEEtran}
\bibliography{IEEEabrv,jddc}
%

 

\ifthenelse{\boolean{CLASSOPTIONdraftcls}}
{}
{
\begin{IEEEbiography}[{\includegraphics[width=1in,height=1.25in,clip,keepaspectratio]{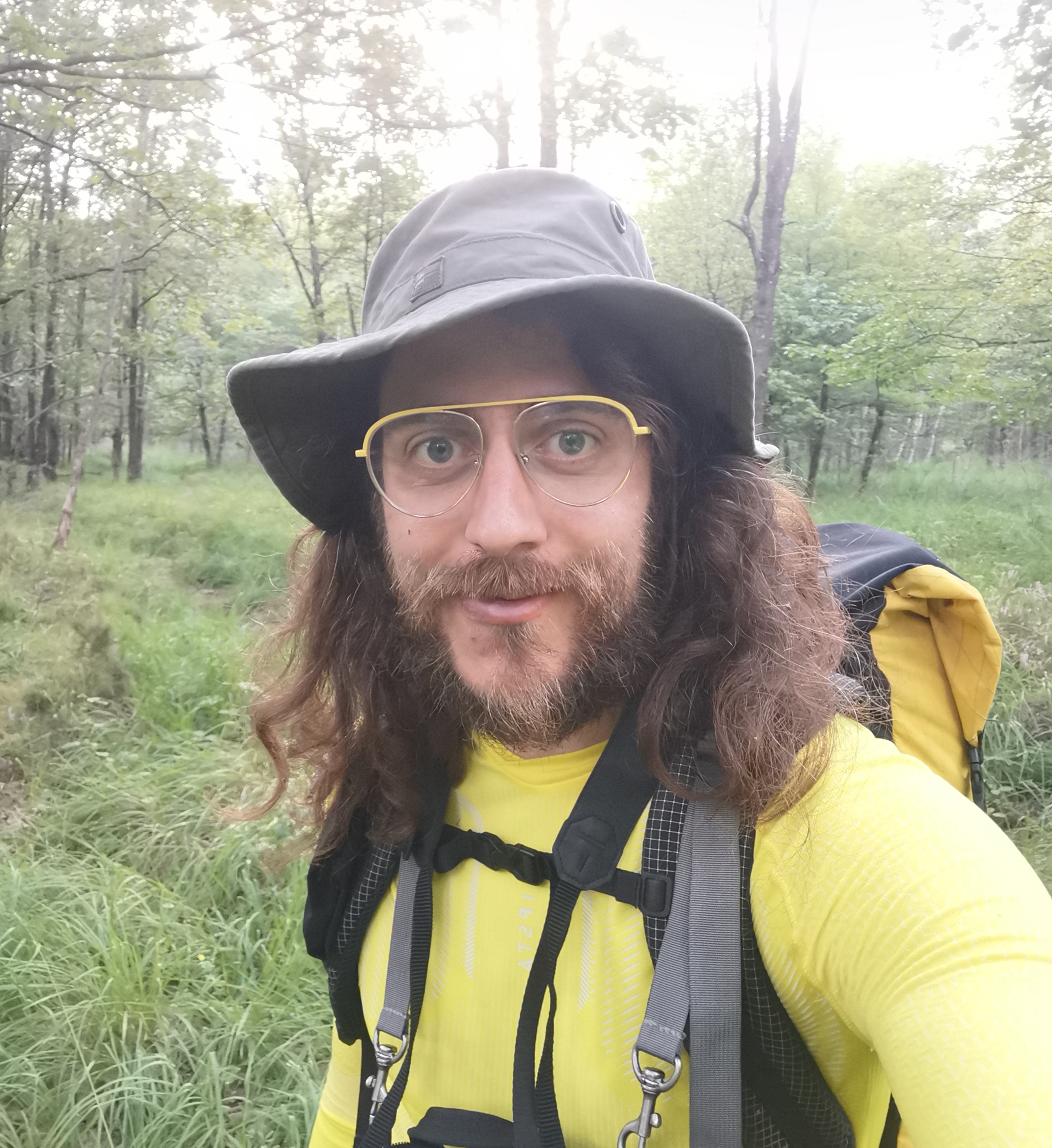}}]{Benoit Brummer}
received a B.Sc. in Computer Engineering from the University of Central Florida and a M.Sc. in Computer Science from the University of Louvain, Louvain-la-Neuve, Belgium. He is currently pursuing a Ph.D. in Joint Learned Compression and Denoising for Improved Image Quality and Reduced Processing Complexity at the Institute for Information and Communication Technologies, Electronics and Applied Mathematics (ICTEAM) at UCLouvain, under the supervision of Professor Christophe De Vleeschouwer.

His research interests include image denoising and processing, optimization problems, and computational efficiency. He has contributed to the development of the Natural Image Noise Dataset. His extracurricular interests include outdoor adventures, photography, forest gardening, nutrition, video and board games, and open-source software.}
\end{IEEEbiography}

\begin{IEEEbiography}[{\includegraphics[width=1in,height=1.25in,clip,keepaspectratio]{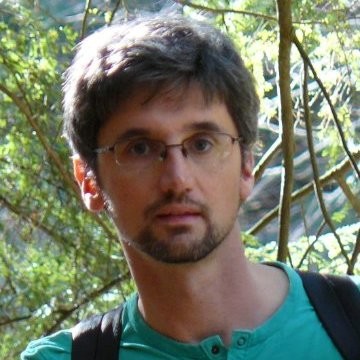}}]{Christophe De Vleeschouwer}
is a Research Director with the Belgian NSF and a \textit{Professeur extraordinaire} with the ISP Group, University of Louvain, Belgium. He was a Senior Research Engineer with IMEC from 1999 to 2000, a Post-Doctoral Research Fellow at the University of California, Berkeley, from 2001 to 2002, and EPFL in 2004, and a Visiting Faculty member at Carnegie Mellon University from 2014 to 2015. He has co-authored over 40 journal papers or book chapters and holds two patents.

His main interests lie in video and image processing for content management, transmission, and interpretation. He is particularly enthusiastic about nonlinear and sparse signal expansion techniques, ensemble classifiers, multiview video processing, and the graph-based formalization of vision problems. He served as an Associate Editor of the \textit{IEEE Transactions on Multimedia} and has been a Technical Program Committee Member for numerous conferences focused on video communication and image processing.
\end{IEEEbiography}
}


\vfill

\end{document}